\documentclass[10pt,twocolumn,letterpaper]{article}

\usepackage[pagenumbers]{cvpr} 

\usepackage{graphicx}
\usepackage{amsmath}
\usepackage{amssymb}
\usepackage{booktabs}
\usepackage{multirow}
\usepackage{multicol}
\usepackage{color}
\usepackage{xcolor}
\usepackage{booktabs,tabularx}

\usepackage[pagebackref,breaklinks,colorlinks]{hyperref}

\usepackage[capitalize]{cleveref}
\crefname{section}{Sec.}{Secs.}
\Crefname{section}{Section}{Sections}
\Crefname{table}{Table}{Tables}
\crefname{table}{Tab.}{Tabs.}

\def\cvprPaperID{9479} 
\def\confName{CVPR}
\def\confYear{2023}

\begin{document}


\title{Cross-Modal Synergies: Unveiling the Potential of Motion-Aware Fusion Networks in Handling Dynamic and Static ReID Scenarios}

\author{Fuxi Ling\\
Hangzhou Dianzi University \\
lingfuxi@gmail.com
\and
Hongye Liu\\
China Jiliang University \\
hongyeliu@mail.cjlu.edu.cn
\and
Guoqiang Huang\\
Zhejiang Gongshang University \\
guoqianghuang@mail.zjgsu.edu.cn
\and
Jing Li\\
Zhejiang Gongshang University \\
jingli@mail.zjgsu.edu.cn
\and
Hong Wu\\
Hangzhou Dianzi University\\
wuhong.wh@gmail.com
\and
Zhihao Tang\\
Hangzhou Dianzi University
}
\maketitle

\begin{abstract}
\label{abstract}
Navigating the complexities of person re-identification (ReID) in varied surveillance scenarios, particularly when occlusions occur, poses significant challenges. We introduce an innovative Motion-Aware Fusion (MOTAR-FUSE) network that utilizes motion cues derived from static imagery to significantly enhance ReID capabilities. This network incorporates a dual-input visual adapter capable of processing both images and videos, thereby facilitating more effective feature extraction. A unique aspect of our approach is the integration of a motion consistency task, which empowers the motion-aware transformer to adeptly capture the dynamics of human motion. This technique substantially improves the recognition of features in scenarios where occlusions are prevalent, thereby advancing the ReID process. Our comprehensive evaluations across multiple ReID benchmarks, including holistic, occluded, and video-based scenarios, demonstrate that our MOTAR-FUSE network achieves superior performance compared to existing approaches.
\end{abstract}

\section{Introduction}
\label{sec:intro}

In contemporary urban landscapes, the surge in population and infrastructural density necessitates sophisticated surveillance solutions that can effectively manage public safety and streamline security operations. Person re-identification (ReID) has become integral to these systems, enabling the identification of individuals across dispersed surveillance cameras with varying perspectives and conditions ~\cite{yang2014salient, liao2015person, zheng2012reidentification, zhang2018robust, zhang2018learning}. The deployment of ReID spans multiple domains from enhancing public safety initiatives to optimizing retail customer experiences, marking its significance in both civil and commercial sectors ~\cite{bai2022salient, hou2019interaction, AP3D, CAViT, zhao2024harmonizing, zhao2024multi}.

The evolution of ReID technology over recent years has been catalyzed by breakthroughs in deep learning, which have substantially refined the accuracy of identifying unobscured subjects. Nevertheless, the application of ReID in natural, uncontrolled settings is frequently hampered by challenges such as inconsistent lighting, physical blockages, and the limited field of view from stationary cameras ~\cite{wang2022feature, sharma2021person}.

The static nature of traditional image-based ReID frameworks often falls short in dynamic urban environments where human activity is fluid and unpredictable. These systems struggle to capture the sequential and behavioral subtleties crucial for recognizing individuals in densely populated areas where visual obstructions are common ~\cite{tang2023character, tang2022few, liu2023spts}.

In response to these limitations, there has been a pivot towards video-based ReID approaches. These methods exploit the rich temporal data inherent in video streams, offering enhanced capabilities in scenarios plagued by frequent occlusions or pronounced variations in individual poses ~\cite{ge2018fd, arnab2021vivit}. By analyzing sequences of movements, video-based systems provide a deeper insight into behavioral patterns, substantially aiding in the distinction of persons in crowded scenes.

Our novel Motion-Aware Fusion (MOTAR-FUSE) network is designed to synthesize the strengths of both static and dynamic analysis techniques. MOTAR-FUSE utilizes inferred motion from static images, a method traditionally employed in video processing, to augment the identification process in image-based systems. This synthesis allows for a comprehensive analysis that addresses the typical challenges of occlusions and environmental variability found in urban surveillance settings ~\cite{ViT}.

\begin{figure}[t]
    \begin{center}
        \includegraphics[width=1.0\linewidth] {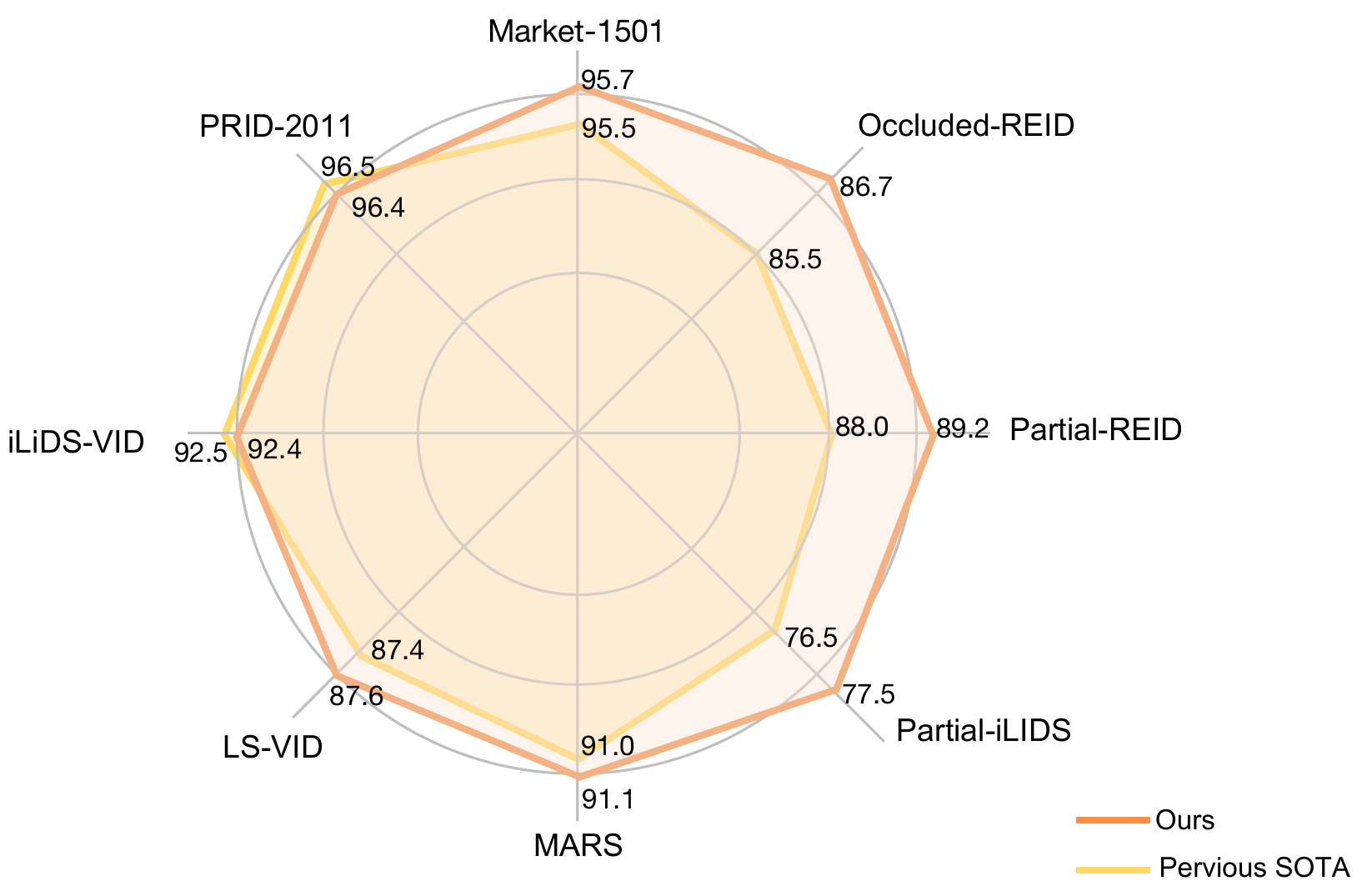}
    \end{center}
    \caption{Graphical depiction of MOTAR-FUSE's comparative advantage over standard ReID models in diverse operational contexts.}
    \label{intro}
\end{figure}

This transformative approach not only remedies the deficits of previous systems but also adapts seamlessly to the complexities of real-world surveillance. Extensive testing across varied ReID benchmarks underscores MOTAR-FUSE's unparalleled efficacy, particularly in environments characterized by intricate occlusions and fluctuating conditions. Our findings affirm MOTAR-FUSE’s potential to redefine the operational capabilities of urban security systems.

\section{Related Work}

\label{sec:related}

This section outlines key developments in the domain of person re-identification (ReID), underscoring major advancements in image-based identification and the innovative use of motion data through advanced segmentation techniques, which have significantly shaped our research approach.

\subsection{Image Person Re-Identification}

Image-based person ReID, covering both holistic and partial visibility scenarios, focuses on the consistent identification of individuals from different camera perspectives. The expansion of comprehensive datasets ~\cite{BiCnet-TKS, tang2024mtvqa, shan2024mctbench, wang2024pargo,tang2024textsquare} combined with breakthroughs in deep learning, particularly through the use of transformer technologies, has considerably advanced our capability to extract nuanced human features. This advancement has set new standards in holistic ReID tasks. For example, TransReID~\cite{he2021transreid,liu2024rethink} integrates a transformer with unique modules like a jigsaw patch mechanism and side information embeddings, thereby enhancing performance significantly. Efforts have also been concentrated on overcoming challenges related to occlusion. For instance, PAT~\cite{li2021diverse} utilizes a novel encoder-decoder structure with adaptable part prototypes dedicated to occluded ReID situations, achieving noteworthy outcomes. Moreover, FED~\cite{wang2022feature} targets improvements against obstructions with strategic data augmentation and introduces mechanisms for erasing occlusions and diffusing features, enhancing the clarity of pedestrian identification. DPM~\cite{tan2022dynamic} leverages a hierarchical mask generator to enhance the overall prototype and maintain a comprehensive imagery dataset, facilitating seamless alignment without external aids. Despite progress, the efficacy of these models is often limited by unexpected occlusions and predominantly relies on visual cues from subjects \cite{wang2024pargo, sun2024attentive, lu2024bounding, zhao2024tabpedia}.

\subsection{Video Person Re-Identification}

Video-based ReID utilizes temporal data to address common issues like occlusion and motion blur found in static-image methods. This modality gains from immediate access to motion analytics and optical flow, thereby improving identification precision. Prominent methods involve temporal attention mechanisms that highlight important frames and disregard those of lower quality, and the use of self-attention or Graph Convolutional Networks (GCNs) ~\cite{gao2020pose, tang2022optimal, tang2022youcan, feng2023unidoc} to bolster temporal linkages and dependencies among frames. A particular methodology~\cite{yin2020fine, feng2024docpedia} merges an RNN-mask framework with a pre-trained keypoint detector to derive elaborate motion and local part features. Another distinct strategy, the mutual attention network~\cite{kiran2021flow}, applies optical flow in a Siamese network setting to extract essential spatiotemporal features for ReID. SBM~\cite{bai2022salient} introduces a temporal relation-based method that enhances differential analyses for richer representations, although such approaches often face challenges with the integration of global-range features and are computationally intensive.

\subsection{Motion-Guided Segmentation}

In the area of motion-guided segmentation, Siarohin \emph{et al.}~\cite{siarohin2021motion} have innovated a self-supervised learning strategy that leverages motion details for effective segmentation of human parts. This methodology, drawing from foundational efforts~\cite{jakab2018unsupervised, zheng2019pose, siarohin2019first}, uses a reconstruction objective to dissect semantic and appearance attributes separately. While this is resource-intensive, it furnishes critical insights into how motion data can distinguish human elements and provide latent motion indicators, profoundly influencing our approach.

Distinctly, our methodology efficiently handles both image- and video-based ReID tasks with a novel motion-aware transformer that directly extracts motion data from static images, enhancing the feature set for ReID. This dual capability underscores our method’s innovation, distinguishing it from conventional strategies in the field.

\begin{figure*}[t]
    \begin{center}
    	\includegraphics[width=1.0\linewidth] {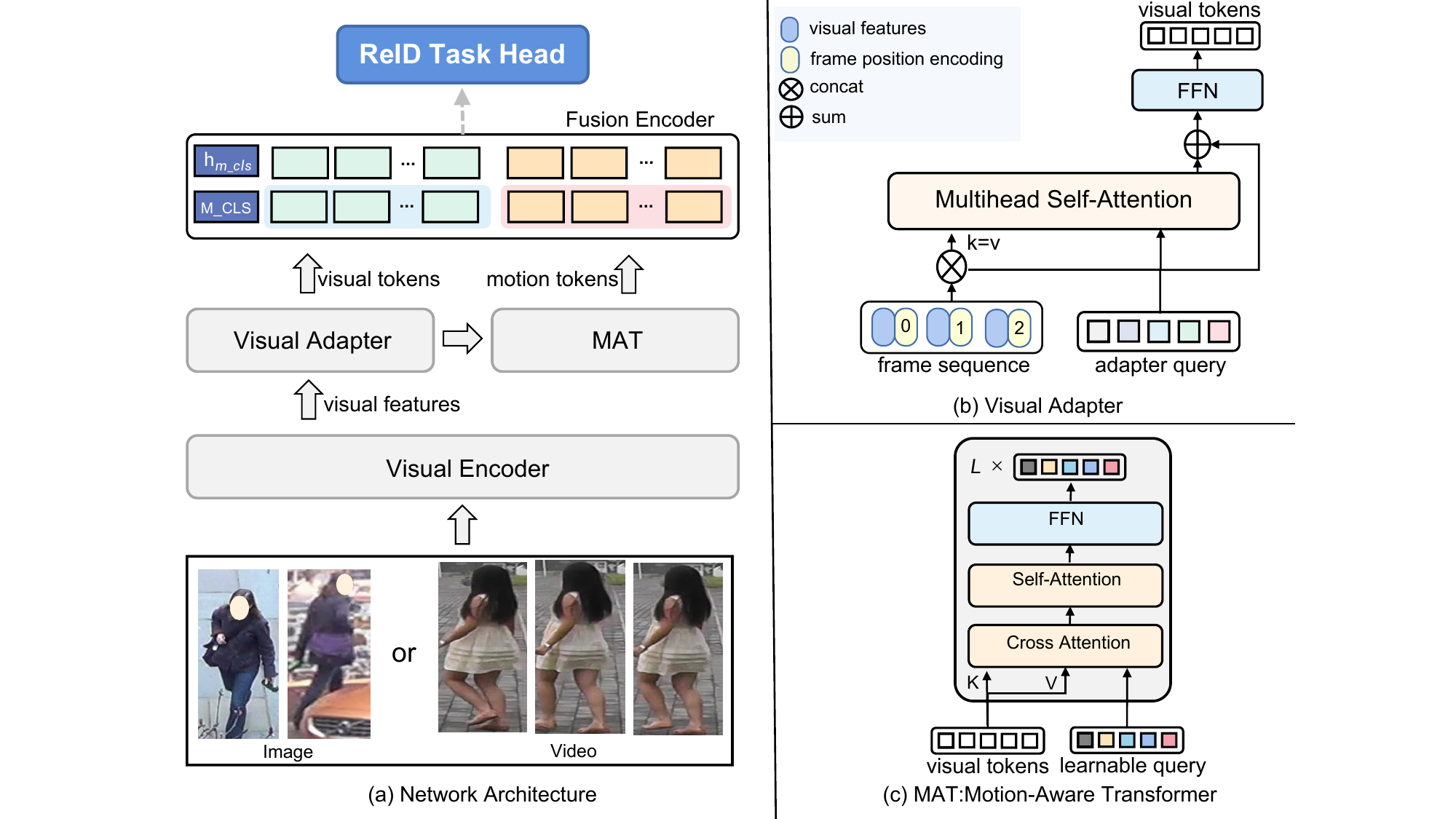}
    \end{center}
    \caption{An overview of the proposed MOTAR-FUSE module.}
    \label{pipeline}
\end{figure*}

\section{Methodology}
\label{sec:method}
Our Motion-Aware ReID system is engineered to exploit spatial and temporal data derived from static images, thereby significantly enhancing the accuracy of person re-identification. The system architecture, illustrated on the left side of Fig.~\ref{pipeline}, integrates several key components: a \textbf{visual encoder}, a \textbf{visual adapter}, a \textbf{motion-aware module}, and a \textbf{fusion encoder}.

The visual encoder initiates the process by capturing detailed spatial features from the entire image. These features are subsequently converted into a series of visual tokens by the visual adapter. The motion-aware module processes these tokens to extract dynamic motion tokens, which, along with a hybrid class token, feed into the fusion encoder producing the final output used in pedestrian identification.

\subsection{Visual Encoder \& Visual Adapter}
\label{sec:architecture}
Refer to Fig.~\ref{pipeline} (b) for a diagram of this component. Utilizing a pre-trained Vision Transformer (ViT-B/16)~\cite{ViT}, the visual encoder outperforms conventional convolutional networks by analyzing complex image backgrounds more effectively. Training involves pairing images within each batch to compute motion consistency loss, which is detailed in Section~\ref{sec:Motion-Aware Transformer}. The images are dissected into patches, linearly projected to a uniform vector size across all layers, and enhanced with positional embeddings to enrich the visual features.

The visual adapter then condenses these enhanced features into visual tokens, optimizing computational efficiency while retaining essential spatio-temporal information. This module's adaptability to various media inputs is achieved by adjusting the frame count and modifying positional embeddings to suit different input types.

\subsection{Motion-Aware Transformer}
\label{sec:Motion-Aware Transformer}
Illustrated in Fig.~\ref{pipeline}-(c), this module transforms visual tokens into motion tokens through a sequence of operations. A cross-attention layer initially isolates specific human body parts from the visual tokens, which are then refined into motion tokens by subsequent self-attention and feed-forward layers, capturing essential dynamic information. We introduce a specialized motion consistency task with a custom loss function to ensure the transformer's efficacy during training, particularly effective when processing single-image inputs at inference.

\subsection{Fusion Encoder}
\label{Fusion Encoder}
This encoder amalgamates visual and motion tokens into a unified representation, encapsulated in a hybrid class token. This comprehensive token, processed through an advanced transformer-based encoder, captures an integrated representation of human features crucial for effective ReID. The encoder outputs an additional token vital for the final identification tasks, integrating linear projections of both token types and enabling their integration through sophisticated cross-attention mechanisms.

\subsection{Training and Inference}
End-to-end training of the system begins with a stabilization phase using video datasets to solidify motion token generation. The model is refined using a hybrid loss function that combines cross-entropy and triplet loss, optimizing the system for precise pedestrian identification.

During inference, the output from the fusion encoder, particularly the hybrid class token, serves as the primary feature set for retrieval tasks, ensuring accurate and efficient person re-identification across diverse scenarios. This method maximizes the utilization of spatial and temporal data inherent in human movements, as expressed by the following equations:

\begin{equation}
\label{loss}
    \mathcal{L} = \lambda_{g}\mathcal{L}_{g} + \lambda_{mc}\mathcal{L}_{mc},
\end{equation}
where $\lambda_{g}$ and $\lambda_{mc}$ are scaling factors for the cross-entropy loss $\mathcal{L}_{g}$ and motion consistency loss $\mathcal{L}_{mc}$ respectively:

\begin{equation}
    \mathcal{L}_{g} = \mathcal{L}_{c}(h_{m\_cls}) + \mathcal{L}_{t}(h_{m\_cls}),
\end{equation}
where $h_{m\_cls}$ are the outputs of the fusion encoder, $\mathcal{L}_{c}$ represents the cross-entropy loss, and $\mathcal{L}_{t}$ represents the triplet loss.

\begin{table*}[t]
		\centering
		\small
		\setlength\tabcolsep{13pt}
		\renewcommand\arraystretch{1}
		\caption{Performance comparison with state-of-the-arts on MARS, LS-VID, iLiDS-VID, and PRID-2011 datasets. Our method achieves competitive performance on four datasets.}
        \begin{tabularx}{1\linewidth}{c|cccccccc}
        \hline
        \multirow{2}*{Method}& \multicolumn{2}{c}{MARS} & \multicolumn{2}{c}{LS-VID} & \multicolumn{2}{c}{iLiDS-VID} & \multicolumn{2}{c}{PRID-2011} \\ 
			~& Rank-1 & mAP& Rank-1 & mAP& Rank-1 & Rank-5& Rank-1 & Rank-5\\ \hline\hline
		M3D~\cite{M3D} & 84.4 & 74.1 & 57.7 & 40.1 & 74.0 & 94.3 & 94.4 & \textbf{100.0} \\
		AP3D~\cite{AP3D} & 90.1 & 85.1 & 84.5 & 73.2 & 88.7 &- & - & - \\
		GRL~\cite{GRL} & 91.0 & 84.8 &- & - & 90.4 & 98.3 & 96.2 & 99.7 \\
		GLTR~\cite{GLTR} & 87.0 & 78.5 & 63.1 & 44.3 & 86.0 & 98.0 & 95.5 & \textbf{100.0} \\
 		MGH~\cite{MGH} & 90.0 & 85.8 & - & - & 85.6 &97.1 & 94.8 & 99.3\\
 		MG-RAFA~\cite{MG-RAFA} & 88.8 & 85.9 & - & - & 88.6 & 98.0 & 95.9 &99.7 \\

		STMN~\cite{STMN} & 90.5 & 84.5 &82.1 & 69.2 &91.5 & - & - &- \\
		CAViT~\cite{CAViT} & 90.8 & 87.2 & 89.2 & 79.2 &93.3 &\textbf{98.0} &95.5 &98.9 \\
		SINet~\cite{SINet}& 91.0 & 86.2 & 87.4 & 79.6 & \textbf{92.5} & - & \textbf{96.5} & - \\
		\hline
		\textbf{Ours} &\textbf{91.1} &\textbf{86.4} &\textbf{87.6} &\textbf{79.9} &92.4 &97.9 &96.4 &\textbf{100.0} \\
		\hline
        
		\end{tabularx}	
		\label{tab:video}
	\end{table*}

\section{Experiments}
\label{Experiments}
This section delineates the empirical validation of our Motion-Aware ReID framework. We outline the setup, engage in a comprehensive evaluation using several benchmark datasets for image and video ReID, and conclude with detailed ablation studies to assess the impact of specific system components.

\subsection{Datasets and Evaluation Metrics}

Our evaluation leverages a variety of datasets designed to challenge and verify the robustness of ReID systems under diverse conditions:
- \textbf{Market-1501}~\cite{zheng2015scalable} is employed to assess our model's performance in environments with minimal occlusions.
- For scenarios with partial visibility, we utilize \textbf{Partial-iLIDS}~\cite{zheng2011person} and \textbf{Partial REID}~\cite{zheng2015partial}, which include images with manually occluded regions.
- \textbf{Occluded REID}~\cite{occluded-reid} tests our model's efficacy in handling severe occlusions.
- Video-based datasets such as \textbf{MARS}~\cite{MARS}, \textbf{LS-VID}~\cite{LS-VID}, \textbf{iLiDS-VID}~\cite{iLiDS-VID}, and \textbf{PRID-2011}~\cite{PRID-2011} provide dynamic sequences that are crucial for evaluating temporal attribute handling.

Performance metrics include Cumulative Matching Characteristic (CMC) curves and mean average precision (mAP), which offer insights into the ranking accuracy and retrieval quality of our model.

\subsection{Implementation Details}

For consistency, all processed images and video frames are resized to $256\times128$. The network utilizes 16x16 patches. Training parameters include a batch size of 32 clips per batch for video data, accommodating 8 identities, and similar configurations for image data. The system is optimized over 120 epochs using the Adam optimizer, with an added emphasis on data augmentation techniques such as random flipping and erasing to enhance model robustness.

\subsection{Comparative Analysis}

Our method is compared with several state-of-the-art approaches across different datasets, indicating its competitive edge, particularly in handling complex scenarios like occlusions and dynamically captured sequences.

\textbf{Performance on Video-based and Holistic Datasets:}
Our evaluations on datasets like MARS and LS-VID demonstrate superior performance, particularly in terms of Rank-1 accuracy and mAP, underscoring the effectiveness of the visual adapter in distilling spatio-temporal features.

\begin{table}[h]
    \centering
    \caption{Comparison of performance metrics on Market-1501 with state-of-the-art methods.}
    \begin{tabular}{c|cc}
        \hline
        Method & Rank-1 & mAP \\
        \hline\hline
        FPR~\cite{he2019foreground} & 95.4 & 86.6 \\
        PCB~\cite{sun2018beyond} & 92.3 & 77.4 \\
        PGFA~\cite{occluded-duke} & 91.2 & 76.8 \\
        VPM~\cite{sun2019perceive} & 93.0 & 80.8 \\
        HOReID~\cite{wang2020high} & 94.2 & 84.9 \\
        ISP~\cite{zhu2020identity} & 95.3 & 88.6 \\
        PAT~\cite{li2021diverse} & 95.4 & 88.0 \\
        TransReID~\cite{he2021transreid} & 95.0 & 88.2 \\
        FED~\cite{wang2022feature} & 95.0 & 86.3 \\
        DPM~\cite{tan2022dynamic} & 95.5 & 89.7 \\
        \textbf{Ours} & \textbf{95.7} & \textbf{89.8} \\
        \hline
    \end{tabular}
    \label{tab:market}
\end{table}

\begin{table}[t]
		\centering
		\small
		\setlength\tabcolsep{20pt}
		\renewcommand\arraystretch{1}
		\caption{Performance comparison with other methods on occluded-REID dataset.}
		\begin{tabularx}{1\linewidth}{c|cc}
			\hline
			\multirow{2}*{Method}& \multicolumn{2}{c}{Occluded-REID}  \\ 
			~& Rank-1 & mAP\\ \hline\hline
			FPR~\cite{he2019foreground}& 78.3 & 68.0\\
			PCB~\cite{sun2018beyond}& 41.3 & 38.9 \\
			AMC+SWM~\cite{zheng2015partial}& 31.2 & 27.3 \\
			PVPM~\cite{gao2020pose}&70.4& 61.2 \\
			HOReID~\cite{wang2020high}& 80.3 & 70.2 \\
			PAT~\cite{li2021diverse}& 81.6 & 72.1 \\ 
			TransReID~\cite{he2021transreid} &70.2&67.3\\ 
			DPM~\cite{tan2022dynamic} &85.5 &79.7\\
			\hline
			\textbf{Ours}& \textbf{86.7} & \textbf{81.2} \\ 
			\hline
		\end{tabularx}	
		\label{tab:performance_occluded}
	\end{table}
	
\begin{table}[t]
	\centering
	\small
	\setlength\tabcolsep{7pt}
	\renewcommand\arraystretch{1}
	\caption{Performance comparison with other methods on Partial REID dataset, and Partial-iLIDS dataset.}
	\begin{tabularx}{1\linewidth}{c|cccc}
		\hline
		\multirow{2}*{Method} &\multicolumn{2}{c}{Partial-REID} & \multicolumn{2}{c}{Partial-iLIDS} \\ 
		~&Rank-1 & Rank-3 & Rank-1 & Rank-3\\ \hline\hline
		SFR~\cite{he2018recognizing}&56.9 & 78.5 & 63.9 & 74.8 \\
		FPR~\cite{he2019foreground}& 81.0 & - & 68.1 & -  \\
		AMC+SWM~\cite{zheng2015partial}& 37.3 & 46.0 & 21.0 & 32.8\\
		PVPM~\cite{gao2020pose}& 78.3 &87.7&-& - \\
		PGFA~\cite{occluded-duke}&68.0 & 80.0 & 69.1 & 80.9 \\
		VPM~\cite{sun2019perceive}& 67.7&81.9&65.5&74.8 \\
		HOReID~\cite{wang2020high}&85.3 & 91.0 & 72.6 & 86.4 \\
		PAT~\cite{li2021diverse}&88.0 & 92.3 & 76.5 & 88.2\\ 
	    FED~\cite{wang2022feature} &84.6 &- &- &- \\
	    \hline
		\textbf{Ours}& \textbf{89.2} & \textbf{93.2} & \textbf{77.5} & \textbf{89.6}  \\ 
		\hline
	\end{tabularx}	
	\label{tab:performance_part}
\end{table}

\begin{table}[t]
		\centering
		\setlength\tabcolsep{6pt}
		\renewcommand\arraystretch{1.2}
		\caption{The experiments of motion consistency task on Market-1501, Occluded-REID, Partial-REID and MARS. Here, O-REID is the abbreviation of Occluded-REID and P-REID is the abbreviation of  Partial-REID.}
		\begin{tabularx}{1.0\linewidth}{c|cccc}
			\hline
			\multirow{2}*{Method} & \multicolumn{1}{c}{Market-1501} & \multicolumn{1}{c}{O-REID} &\multicolumn{1}{c}{P-REID} &\multicolumn{1}{c}{MARS}\\ 
			~& Rank-1 & Rank-1 & Rank-1 & Rank-1\\ \hline\hline
			w & \textbf{95.7} & \textbf{86.7}& \textbf{89.2} & \textbf{91.1} \\
			w/o & 94.9 & 83.9& 83.6 & 90.5 \\ \hline
		\end{tabularx}	
		\label{tab:motion-information}
\end{table}

\begin{table}[h]
		\centering
		\setlength\tabcolsep{6pt}
		\renewcommand\arraystretch{1.0}
		\caption{The experiments on the length of learnable queries in MOTAR-FUSE on different benchmarks. }
		\begin{tabularx}{1.0\linewidth}{c|cccc}
			\hline
			\multirow{2}*{Length} & \multicolumn{1}{c}{Market-1501} & \multicolumn{1}{c}{O-REID} &\multicolumn{1}{c}{P-REID} &\multicolumn{1}{c}{MARS}\\ 
			~& Rank-1 & Rank-1 & Rank-1 & Rank-1\\ \hline\hline
			2 & 94.0 & 84.0 & 83.7 & 89.9 \\
			4 & 94.4 & 84.3 & 84.2 & 90.1 \\
			6 & 95.0 & 85.7 & 86.3 & 90.4 \\
			8 & 95.2 & 86.3 & 88.9 & 90.9 \\
			10 & \textbf{95.7} & \textbf{86.7} & \textbf{89.2} & \textbf{91.1} \\
			16 & 95.4 & 85.2 & 88.1 & 90.9 \\
			32 & 95.0 & 84.1 & 84.0 & 90.5 \\ \hline
		\end{tabularx}	
		\label{tab:length}
\end{table}

\subsection{Ablation Studies}

To dissect the contributions of specific components, we perform several ablation studies:
- \textbf{Effect of the Motion Consistency Task:} Removing this component significantly diminishes performance on occluded datasets, reinforcing its necessity for handling complex occlusions effectively.
- \textbf{Impact of Pre-training on Video Datasets:} Pre-training enhances the motion-aware transformer's effectiveness, providing it with vital motion insights essential for robust performance across static and dynamic scenes.

These experiments underline the strengths and adaptive capabilities of our ReID system, particularly its proficiency in handling real-world challenges posed by occlusions and dynamic environments.

\begin{table}[t]
	\centering
	\setlength\tabcolsep{4.5pt}
	\renewcommand\arraystretch{1.2}
	\caption{The experiments on whether pre-trained on video dataset.Here, O-REID is the abbreviation of Occluded-REID and P-REID is the abbreviation of  Partial-REID. }
	\begin{tabularx}{1.0\linewidth}{c|cccc}
	\hline
	\multirow{2}*{Method} & \multicolumn{2}{c}{O-REID} & \multicolumn{2}{c}{P-REID}\\ 
	~& Rank-1 & mAP & Rank-1 & Rank-3\\ \hline\hline
	w & \textbf{86.7} & \textbf{81.2} & \textbf{89.2} & \textbf{93.2}\\
	w/o & 86.1 \textcolor{gray}{(-0.6)} & 79.8\textcolor{gray}{(-0.4)} & 88.5\textcolor{gray}{(-0.7)} &92.7\textcolor{gray}{(-0.5)}  \\ \hline
	\end{tabularx}	
	\label{tab:pretrain}
\end{table}

\subsection{Detailed Analysis of Learnable Query Length in Motion-Aware Transformer}
\label{sec:query_length_experiments}
The configuration of learnable queries within the motion-aware transformer is crucial as it directly influences the model's ability to dissect and analyze the human body into distinct parts. Our experiments investigate the impact of varying the query length on the model's performance across several ReID benchmarks, providing insights into the optimal structuring of these queries.

\subsubsection{Experimental Setup and Findings}

We systematically varied the length of the learnable queries in the transformer and assessed the model's performance on datasets like Market-1501, Occluded-REID, Partial-REID, and MARS. The purpose was to determine how the granularity of body part segmentation affects overall identification accuracy.

\textbf{Results Discussion:}
\begin{itemize}
    \item The optimal performance, as evidenced by the highest Rank-1 accuracy, was consistently observed at a query length of 10. This suggests a balanced level of detail that aids effective person re-identification without overwhelming the model with excessive granularity.
    \item When the query length exceeded 10, a notable decrease in performance was observed, particularly on datasets characterized by occlusion and partial visibility (see Tab.~\ref{tab:length}). This decline indicates the potential pitfalls of over-segmentation, where the model may become prone to errors due to an overly fragmented representation of the human figure.
    \item In contrast, the impact on datasets focused on holistic ReID tasks was less pronounced, underscoring the reliance on global features under less challenging conditions.
\end{itemize}

These observations underscore the dual functionality of the fusion encoder, which adeptly integrates both local and global features, enhancing the robustness of the resulting feature representation \( h_{m\_cls} \).

\subsubsection{Comparative Analysis with Previous Methods}

In reviewing earlier human part-based approaches, a common limitation is their reliance on part-aware masks that may inadvertently emphasize background elements, due to high confidence allocations to non-crucial areas. Such designs, while effective under certain conditions, exhibit vulnerabilities in complex occluded environments.

\textbf{Our Approach:}
By integrating motion information derived from static images through a sophisticated motion consistency task, our model extends beyond mere appearance features to incorporate dynamic aspects. This integration significantly enhances the model's resilience and adaptability, enabling robust performance even in scenarios rife with occlusions.

These experiments validate the superiority of a balanced approach to part segmentation within our motion-aware transformer, particularly emphasizing the importance of motion information in enhancing the discriminative power of the learned features, crucial for tackling complex real-world ReID challenges.

\subsection{Influence of Pre-training on Video Datasets}
\label{sec:pretrain_experiments}
One pivotal aspect of our methodology involves the utilization of pre-training on video datasets to enhance the motion-aware transformer's ability to process static images. This section evaluates the impact of this pre-training on the overall performance of our ReID system.

\subsubsection{Effect of Video Pre-training}
Pre-training the motion-aware transformer on video data is intended to endow the model with preliminary motion insights, which are crucial for handling static images effectively. As demonstrated in Table~\ref{tab:pretrain}, omitting this pre-training phase results in a slight but noticeable decrease in performance, specifically a reduction of approximately 0.6 to 0.7\% in Rank-1 accuracy and about 0.5\% in mean Average Precision (mAP). This shows the significant role of motion-derived knowledge in enhancing the predictive prowess of our model.

Although the visual adapter significantly aids in acquiring motion knowledge from video data, it's crucial to note that our system is robust enough to be trained from scratch if necessary. However, in the absence of video pre-training, the model requires more epochs to reach a comparable level of performance, highlighting the efficiency gains provided by this preliminary step.

\subsubsection{Visualization and Analysis of Human Part Segmentation}
\label{sec:visualization}
To further illustrate the practical benefits of integrating motion information, we provide visualizations of human part segmentation achieved through the motion consistency task.

\begin{figure*}[tp]
    \begin{center}
        \includegraphics[width=0.9\linewidth]{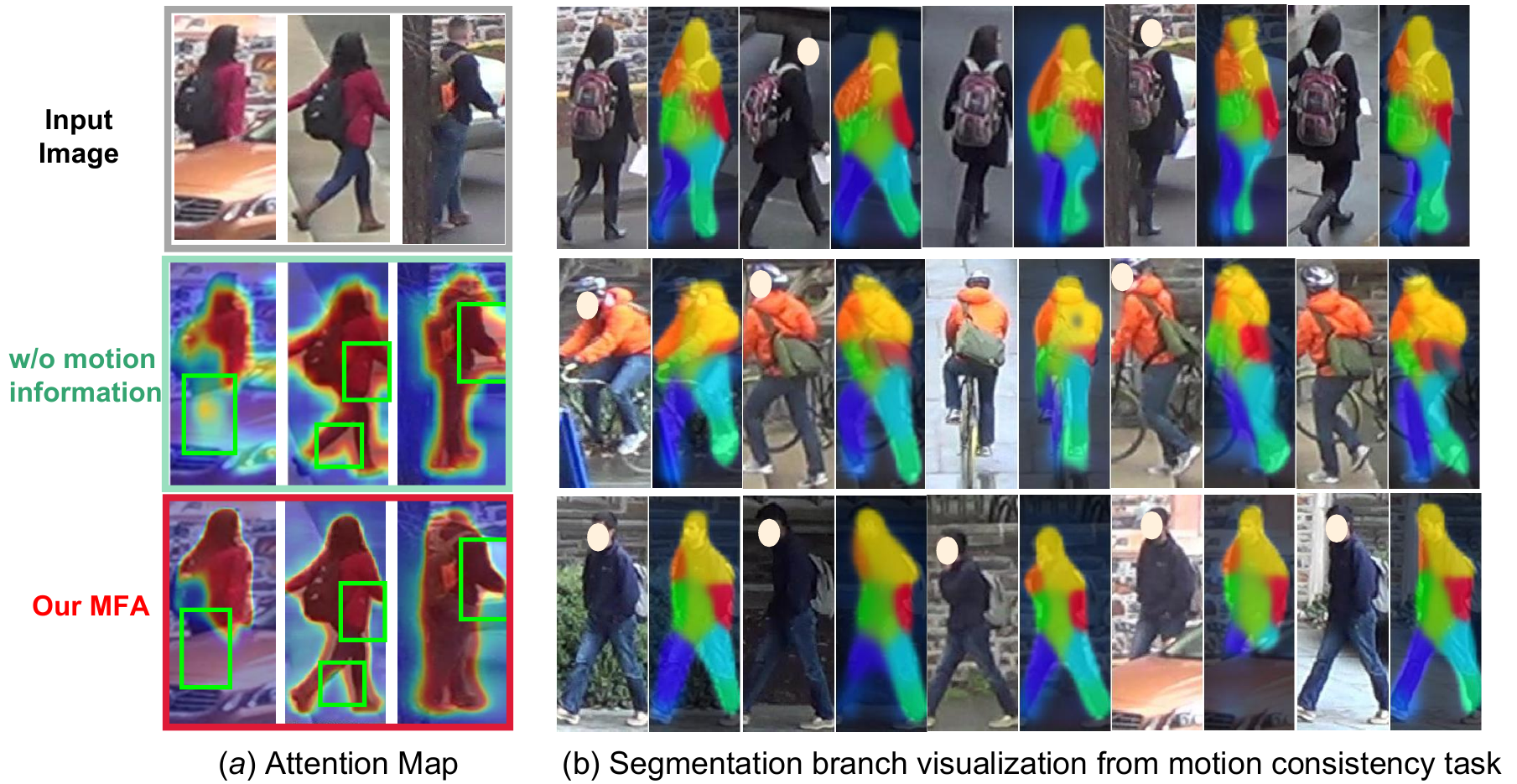}
    \end{center}
    \caption{Visualization of human part segmentation under different conditions: (a) shows enhanced focus on the human body in a cluttered scene; (b) demonstrates robust segmentation when the subject is partially occluded by a bike.}
    \label{visualization}
\end{figure*}

Figure~\ref{visualization} reveals how the addition of motion information refines the attention mechanism of our model:
- \textbf{Fig.~\ref{visualization}-(a)} presents an attention map where the model, informed by motion cues, concentrates more effectively on the human body, even in complex environments. This enhancement is crucial for improving resistance to occlusions and background noise.
- \textbf{Fig.~\ref{visualization}-(b)} depicts a scenario where a pedestrian is occluded by a bicycle. Despite the occlusion, our model effectively discriminates between the human and non-human elements, maintaining accurate segmentation and identification.

These visualizations underscore the effectiveness of motion information in refining the model's ability to handle real-world variability in human appearances, particularly in occluded scenarios. By learning to focus on dynamic human parts, the system demonstrates superior adaptability and accuracy, proving the value of motion data in static image ReID tasks.

\section{Conclusion and Future Work}
\label{sec:conclusion}

This study has significantly advanced the integration of motion information in person re-identification by developing an innovative motion-aware fusion network. Our system stands out by proficiently processing both image and video data, facilitated by a versatile visual adapter. This dual capability enhances the applicability of our model across a variety of real-world scenarios, bridging the gap between static image analysis and dynamic video processing.

Our approach extends the extraction of motion information to still images, a notable enhancement over traditional methods that rely solely on video data. By enabling the fusion encoder to intricately merge visual features with derived motion data, we provide a richer, more nuanced representation of subjects. This enhanced feature integration has led to the establishment of new benchmarks on several ReID datasets, showcasing superior results in both holistic views and occluded scenarios, alongside robust performance in video-based evaluations.

\subsection{Achievements}
The successful implementation of our model has demonstrated its efficacy in various testing environments, achieving state-of-the-art results on well-regarded benchmarks such as Market-1501, Occluded REID, and MARS. These achievements underscore the effectiveness of our method in addressing some of the most challenging aspects of person re-identification, including handling occlusions and extracting useful information from static images.

\subsection{Limitations and Future Directions}
Despite these successes, our approach does have limitations, notably the reliance on extensive pre-training for the motion-aware transformer. This requirement can introduce additional computational overhead and extend the training duration, potentially impacting the scalability and efficiency of deployment in operational settings.

Future research will aim to refine the extraction and integration of spatio-temporal information, to reduce or eliminate the dependence on pre-training phases. Innovations could include the development of more sophisticated motion detection algorithms that require less historical data to train effectively or the implementation of novel neural network architectures that are inherently more adept at recognizing and interpreting motion patterns.

Additionally, we plan to explore the application of our methodology to emerging areas within ReID, such as cross-domain adaptation and low-resource environments, where the ability to generalize from limited or varying data is crucial.

\subsection{Continued Impact}
As we continue our research, we are committed to pushing the boundaries of person re-identification technology. By addressing the current limitations and exploring new applications, we aim to further enhance the robustness, accuracy, and practicality of ReID systems. Our ongoing efforts will focus on making these advanced technologies more accessible and effective in diverse and challenging environments.

In conclusion, this study not only sets a new standard in the field of person re-identification through the innovative use of motion data but also paves the way for future advancements that could revolutionize how these systems are developed and deployed in real-world scenarios.

\clearpage
{\small
\bibliographystyle{ieee_fullname}
\bibliography{egbib}
}
\end{document}